\documentclass[10pt,twocolumn,letterpaper]{article}

\usepackage[dvipsnames]{xcolor}
\usepackage{cvpr} 
\usepackage[accsupp]{axessibility} 
\usepackage{color}
\usepackage{tabularray}
\usepackage{makecell}
\usepackage{multirow}

\definecolor{cvprblue}{rgb}{0.21,0.49,0.74}
\usepackage[pagebackref,breaklinks,colorlinks,citecolor=cvprblue]{hyperref}

\def\paperID{7074} 
\def\confName{CVPR}
\def\confYear{2024}

\def\model{FSC}
\def\modelf{Few-point Shape Completion}

\title{\model: \modelf} 

\author{Xianzu Wu$^{1, 3 \ast}$,~Xianfeng Wu$^{1}$\thanks{Equal contribution.},~~Tianyu Luan$^{2}$,~Yajing Bai$^{1}$,~Zhongyuan Lai$^{1}$\thanks{Corresponding author. This work was supported by the National Key Research and Development Program of China (No. 2021YFC3100804), National Natural Science Foundation of China (No. 62106086), China Scholarship Council (No. 201908420105), Key Research and Development Program of Hubei Province (No. 2021BAD004), State Key Laboratory of Precision Blasting, Jianghan University (No. PBSKL2022201), Research Fund of Jianghan University (No. 2022SXZX16), National Undergraduate Training Programs for Innovation and Entrepreneurship (Nos. 202311072004 and 202311072010), and Undergraduate Research Programs of Jianghan University (Nos. 2023zd111 and 2023zd117).},~~Junsong Yuan$^{2}$
\\
$^1$State Key Laboratory of Precision Blasting, Jianghan University, Wuhan, Hubei 430056, China 
\\ $^2$Department of Computer Science and Engineering, University at Buffalo, Buffalo, New York 14260 
\\ $^3$School of Geophysics and Petroleum Resources, Yangtze University, Wuhan, Hubei 430113, China 
\\ \small \{xianfengwu, yajingbai\}@stu.jhun.edu.cn~~~~~  laizhy@jhun.edu.cn~~~~~ 
\{tianyulu, jsyuan\}@buffalo.edu~~~~~ 202007847@yangtzeu.edu.cn
}

\begin{document}
\maketitle

\begin{abstract}

While previous studies have demonstrated successful 3D object shape completion with a sufficient number of points, they often fail in scenarios when a few points, e.g. tens of points, are observed. Surprisingly, via entropy analysis, we find that even a few points, e.g. 64 points, could retain substantial information to help recover the 3D shape of the object. To address the challenge of shape completion with very sparse point clouds, we then propose \modelf{} (\model{}) model, which contains a novel dual-branch feature extractor for handling extremely sparse inputs, coupled with an extensive branch for maximal point utilization with a saliency branch for dynamic importance assignment. This model is further bolstered by a two-stage revision network that refines both the extracted features and the decoder output, enhancing the detail and authenticity of the completed point cloud. Our experiments demonstrate the feasibility of recovering 3D shapes from a few points. The proposed  \modelf{} (\model{}) model outperforms previous methods on both few-point inputs and many-point inputs, and shows good generalizability to different object categories. Code is available at \url{https://github.com/xianzuwu/FSC}.

\end{abstract}

\section{Introduction}

Point-cloud-based shape completion is showing exciting prospects in applications such as autonomous driving, robotics, augmented reality, etc. Limited by the sensors' range and resolution, the point cloud the sensors generate can often be sparse, providing only a few points as input. This application scenario makes few-point shape completion critically important. However, compared to existing works that utilize a few thousand points as input, completing the point cloud with no more than a few dozen points not only faces the absence of local details in the input but more importantly, the highly compromised global shape information. This disparity makes the task of shape completion with such limited input even more challenging.

\begin{figure}[t]
    \centering
    \includegraphics[width=1.0\linewidth]{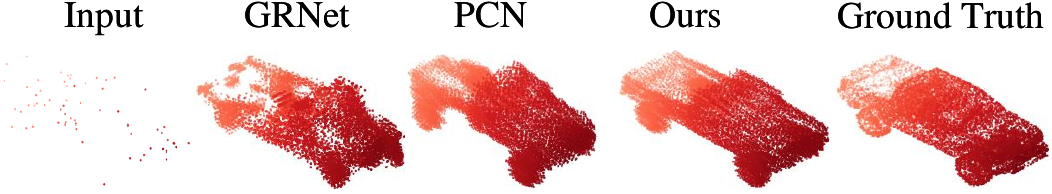}
    \caption{We use 64 points as input for point cloud completion. The PCN~\cite{yuan2018pcn} result does not result in good general shapes. GRNet~\cite{xie2020grnet} result has unexpected holes due to the lack of dense representations. In comparison, our few-point shape completion method can provide much more reasonable results. }	
    \label{fig:teaser}
\end{figure}

Despite the progress in point cloud-based shape completion highlighted in works like~\cite{yuan2018pcn, achlioptas2017learning, Zong2021ASHFNetAS}, few have tackled the challenge of completing a point cloud with few input points. Previous works like \cite{yuan2018pcn,xie2020grnet}  typically utilize a few thousand points as input and demonstrate competent completion capabilities with this volume of input. However, these works do not investigate the potential for point cloud completion when the number of input points is significantly reduced, which scenario might arise in many practical applications. They do not explore the extent of the information contained in a few-point input or attempt to leverage this information to enhance point cloud completion.
Previous studies like PCN~\cite{yuan2018pcn} use feature encoders, often inspired by PointNet~\cite{qi2017pointnet} or PointNet++\cite{qi2017pointnetplusplus}, for capturing input shape features. However, with a limited number of points, where each point is more crucial, the feature extraction must effectively use all points and prioritize their importance for diverse shape information. This ensures the feature quality from few-point inputs is on par with many-point inputs. Conversely, voxel-based methods like GRNet~\cite{xie2020grnet} may succeed in global shape recovery, but they face challenges in creating dense representations from a few points. \cref{fig:teaser} shows 64 points input results in inaccurate global shapes with PCN, while voxel-based approaches GRNet yield sparse outcomes.

To explore how much information that contained in the few-point input, we use Shannon Entropy~\cite{silverman2015cheating} to evaluate the information when we randomly drop the points from input. We found that even when the input has only 64 randomly chosen points, it still contains nearly 50\% of the ground truth shape information. To effectively recover global shape information and generate a comprehensive point cloud from sparse inputs, we designed a dual-branch feature extractor. The extensive branch ensures full utilization of each input point, while the saliency branch dynamically assigns importance to points based on their relevance. This design balances equitable point consideration with the flexibility to adjust point significance. Additionally, our two-stage revision network refines the extracted features and the output from the decoder, improving the output's authenticity and detail for a more accurate point cloud.

Specifically, our \modelf{} (\model{}) model contains two primary modules: recovery and revision. The recovery module employs an MLP-based branch for comprehensive feature extraction, utilizing a simple yet effective MLP-based design to fulfill our objective of extensive feature extraction. In the meantime, the salient feature extraction branch incorporates a transformer-based design, facilitating dynamic exploration of the importance of individual points. For the revision module, we leverage Wasserstein GAN (WGAN) to act upon both the latent feature and the point cloud results, with both WGAN generators being fashioned as simple MLPs. Our approach goes beyond utilizing WGAN solely for output point cloud enhancement; it also ensures that the global shape feature is closely aligned with the ground truth input feature generated from our recovery module, thereby improving the detailed shape of the point cloud. Additionally, we employ a PointNet-based module to decode the extracted feature into finely detailed point clouds.

Our contributions are as follows:
\begin{itemize} 
    \item We investigate the potential of completing point clouds with a few input points. This research gives a minimum of how many points would be enough to complete the entire point cloud.
    \item We present a solution named \modelf{} (\model{}) model capable of completing a point cloud from a few sparse points. To our knowledge, our work is the first one that focuses on the few-point completion task.
    
    \item We design a dual-branch feature recovery architecture to separately capture the extensive and salient information. This dual-branch design equitably utilized the information of all points while adaptively adjusting the importance of each point.
    \item We design a two-stage revision module that uses WGAN directly on the feature space and on point cloud results. This two-stage revision ensures the global shape feature aligns more closely with that of the authentic point cloud while enhancing the shape.
\end{itemize}

Our experiments demonstrate that our method surpasses previous approaches with both many and few input points. We also verify that our method has good robustness to different input point numbers, unseen point cloud categories, and different randomized input points.

\section{Related Work}

\subsection{Point-cloud-based Shape Completion}
Point cloud completion \cite{Pan2020ECGEP, Nie2020skeleton, Zhang2021VGPCC, zhang2020detail, Wang2019DeepCG, Zhu2021TowardsPC, Zhang2021GAN, Wang2021LEG, CHEN2021GeneCGAN, Xin2021Cycle4Completion, Egiazarian2019LatentSpaceLP, Wang2020PointCC, Xie2021SPG, Wang2022PointAttNYO} is a vital task in 3D computer vision with a multitude of applications, such as robot perception, autonomous driving, and augmented reality. Voxel-based and point-based methods are the two primary categories of point cloud completion approaches. The voxel-based method converts the input point cloud into voxels and subsequently generates complementary results in both voxel and point cloud representations. GRNet~\cite{xie2020grnet} exemplifies a voxel-based method that leverages 3D CNNs to extract voxel features and multilayer perceptrons (MLPs) to produce complementary results in point cloud representation. Nonetheless, voxel-based methods exhibit a lack of robustness when confronted with a limited number of input points, leading to missing voxel neighborhoods and, consequently, the generation of results containing holes.

In contrast, point-based methods directly perform point cloud completion on the input points. PCN~\cite{yuan2018pcn}, a trailblazing point-based method, segregates the completion process into an encoder and a decoder. The encoder utilizes PointNet to extract global features, while the decoder employs fully connected and FoldingNet networks to generate coarse and fine point clouds, respectively. Given that the completion process negates the need for neighborhood calculation, point-based methods are more resilient to variations in the number of input points. However, their completion outcomes are often suboptimal, prompting the advent of various enhancement methods such as PF-Net~\cite{huang2020pfnet}, PointWOLF~\cite{kim2021pointwolf}, RSMix~\cite{lee2021rsmix},TopNet~\cite{tchapmi2019topnet} and SA-Net~\cite{wen2020point} . These methods concentrate on procuring more representative point clouds with distinct geometric features and progressively amplifying the resolution of the generated point cloud.

\subsection{Point Cloud Analysis}

Point cloud analysis encompasses a range of tasks including segmentation, object detection, registration, completion, and surface reconstruction, with deep learning-based methods, such as \cite{qi2017pointnet, wang2019dynamic, yang2020pointflownet, luan2024spectrum, luan2021pc, luan2023high,choe2022pointmixer,ma2021pointmlp,qian2022pointnext,chen2023unsupervised,pang2022masked,levi2023epic,ren2022wolfmix,goyal2021simpleview,chen2020pointmixup}, being widely implemented. In these works, point clouds are typically encoded either by leveraging their connections with neighboring points, as seen in methods like PointNet++\cite{qi2017pointnetplusplus}, DGCNN\cite{wang2019dynamic}, and Point Transformer~\cite{Zhao2021proceedings}, or by employing simple networks such as MLP, as in the encoder of PCN~\cite{yuan2018pcn}. While these encoding designs have proven to be effective in previous point cloud completion tasks, they tend to fall short when dealing with inputs comprising a limited number of points. Given this constraint of limited input information, our point cloud encoder must be significantly more effective than those employed in previous point cloud analysis approaches.

\begin{figure*}[t]
	\centering	
	\includegraphics[width=0.8\linewidth]{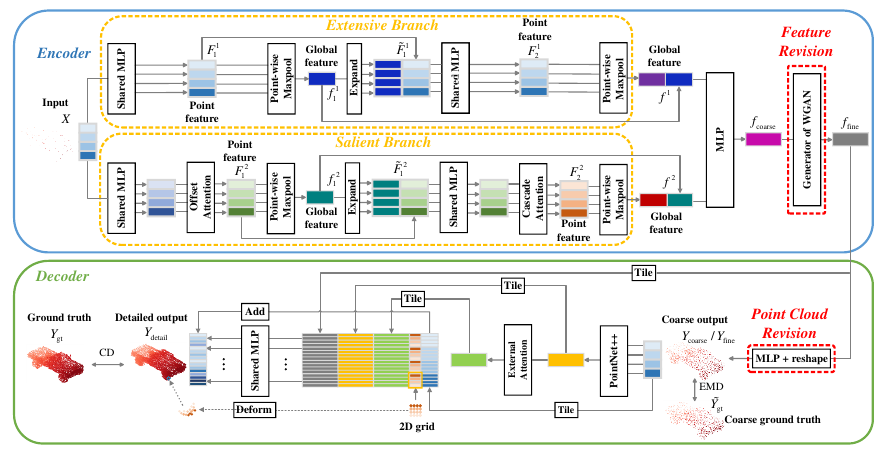}
	\caption{\model{} pipeline. The dual-branch feature extraction network abstracts the input point cloud $X$ as a rough feature vector $f_{\rm{coarse}}$. The two-stage revision network corrects $f_{\rm{coarse}}$ and the coarse point cloud $Y_{\rm{coarse}}$ as $f_{\rm{fine}}$ and $Y_{\rm{fine}}$, respectively. The detail point cloud $Y_{\rm{detail}}$ is generated by using $f_{\rm{fine}}$, $Y_{\rm{fine}}$ and its point features.}	
\label{fig:pipeline}
\end{figure*}

\section{Few-point Information Analysis}
\label{sec:fpfh}
We first analyze how many input points can provide enough information for completing the point cloud. To measure the information, we use Fast Point Feature Histograms (FPFH) Entropy in~\cite{hu2023tree} to encode the shape and calculate its Shannon Entropy. FPFH of a point cloud is defined in~\cite{rusu2009fast}. FPFH is a widely used shape descriptor that encodes the shape of a point cloud into a histogram. Having this quantization, we can calculate the FPFH Entropy as 
\begin{equation}
S = -\sum_i\textit{FPFH}_i\log(\textit{FPFH}_i),
\label{eq:fpfh}
\end{equation}
where $\textit{FPFH}_i$ is the $i$th bin of normalized FPFH. 

Having FPFH entropy, we can calculate how much percentage of information has been lost when we reduce the number of points. We show in \cref{fig:FPFH_Entropy} the information reduction when the input number of points changes. We found that only 64 points input contains almost 50\% of the shape information under a reasonable quantization level.
In the next section, we try to design a solution that can complete the point cloud of 64 or more points.


\section{\modelf{}}
Our task is to design a point cloud completion network architecture that can take very few partial points as input. Specifically, given the vertex locations of the input point cloud $X \in R^{n \times 3}$ ($n$ is the input point number), our network can be abstracted as:
\begin{equation}
  Y = F(X),
  \label{eq:task}
\end{equation}
where $Y \in R^{m\times 3}$ is the output point cloud and $m$ is the output point number. Different from the previous point cloud completion task that takes thousands of points as input, few-point completion would require a much smaller number of points. Typically, we would constrain the point number to no less than 64, \ie{} $m \geq 64$. 
\subsection{Pipeline}

Our \modelf{} (\model{}) model uses an encoder-decoder architecture illustrated in \cref{fig:pipeline}\textcolor{red}{a}. This architecture consists of a dual-branch encoder that transforms the input point cloud into a latent feature space. The dual branches are specifically designed to extract both extensive and salient information from the point cloud. After extracting these features, a two-stage revision network is applied to enhance the feature quality. This revision process is set up adversarially, aiming to bring the point cloud result closer to an actual point cloud and align the features more closely with those derived from a real point cloud. This method improves both the features and the quality of the point cloud results. Subsequently, the revised features are fed into a decoder network to generate a detailed point cloud. The fusion of local and global representations of the point features is achieved using PointNet++\cite{qi2017pointnetplusplus} and multi-head external attention\cite{guo2022beyond}. Finally, the FoldingNet approach~\cite{yang2018Foldingnet} is employed to create the final detailed point cloud.

\subsection{Dual-branch Feature Extraction}

We focus on developing a feature extraction network that can effectively extract the coarse features ${f_{{\rm{coarse}}}}$ of point clouds $X$ as 
\begin{equation}
     f_{\rm{coarse}} = F_{\rm{ext}}(X),
\label{eq:dual_branch}
\end{equation}
Our approach addresses the challenge of sparse feature extraction from point clouds with limited points. We employ a dual-branch feature extraction network, combining point-specific and global information. This includes a double-layer stacking technique for a richer representation and a transformer-based network to enhance global feature correlation. We balance detailed shape features with globally relevant ones, leveraging the strengths of PointNet\cite{qi2017pointnet} and PCN~\cite{yuan2018pcn} methodologies. This results in a robust, effective solution for feature extraction in sparse point clouds.

\textbf{Extensive branch.} The extensive branch of the network has a structure that is similar to the PCN feature extraction network, which also utilizes a two-layer stacking architecture. In the first layer, the multilayer perceptron network takes the $n \times 3$-dimensional sparse point cloud $X$ as input to generate the feature matrix $F_1^1$. Then, the feature vector $f_1^1$ is obtained through maximum pooling. In the second layer, the global features $f_1^1$ are merged into each row of the feature matrix $F_1^1$, resulting in an expanded point feature matrix $\tilde F_1^1$. Subsequently, the feature vector $f_2^1$ is generated by applying two multilayer perceptrons and maximum pooling sequentially. Finally, $f_1^1$ and $f_2^1$ are combined to form the feature vector ${f^1}$ of the first branch.

\textbf{Salient branch.} The structure of the salient branch is similar to that of the first branch. However, it differs in that the second branch includes an offset attention module~\cite{guo2021pct} inserted after the multilayer perception network in the first layer of the two-layer stacked structure. Additionally, a two-layer cascaded attention module~\cite{WU2022108413} is inserted between the two multilayer perception networks in the second layer to enhance the global representation of the point features in the second layer. The goal of this modification is to improve the effectiveness of the feature extraction process for the salient branch. Finally, the coarse feature vector ${f_{{\rm{coarse}}}}$ is obtained by concatenating the feature vectors ${f^1}$ and ${f^2}$ of the two branches together.


\subsection{Two-stage Revision}

The two-stage revision network is used to correct the coarse feature vector ${f_{{\rm{coarse}}}}$ and the rough point cloud ${Y_{{\rm{coarse}}}}$.
As shown in \cref{fig:pipeline}\textcolor{red}, this process can be abstracted as:
\begin{equation}
\begin{split}
     &f_{\rm{fine}} = F_{\rm{rf}}(f_{\rm{coarse}})), \\
     &Y_{\rm{coarse}} = \rm{MLP}(f_{\rm{fine}}), \\
     &Y_{\rm{fine}} = F_{\rm{rp}}(Y_{\rm{coarse}}), 
     \label{eq:two_stage}
\end{split}
\end{equation}
where $f_{\rm{fine}}$ is the revised feature using the first stage feature revision. $Y_{\rm{coarse}}$ is the resulting coarse point cloud getting from the revised feature. $Y_{\rm{fine}}$ is the shape revised point cloud with the same number of points as $Y_{\rm{coarse}}$. $F_{\rm{rf}}$ and $F_{\rm{rp}}$ are the feature and point cloud revision network.
In this process, we aim to generate a point cloud that is closer to the real point cloud and revise the feature to be more similar to the features generated from the real point cloud. This adversarial approach leads to a better quality of both the feature and the point cloud results.


Specifically, our two-stage revision network refines both the coarse feature vector and the rough point cloud. In the first stage, a seven-layer fully connected generator adjusts the coarse feature vector, using feature fusion through layer stitching. An attention-enhanced four-layer discriminator network analyzes feature channel correlations. The second stage focuses on aligning the rough point cloud with the downsampled true point cloud, using a three-layer generator and a four-layer discriminator. This method effectively enhances the fidelity of both features and point clouds, ensuring they closely resemble those from real point clouds.

Finally, the decoder network is used to generate the detail point cloud output ${Y_{{\rm{detail}}}}$ from the shape revised point cloud ${Y_{{\rm{fine}}}}$ as 
\begin{equation}
     Y_{\rm{detail}} = F_{\rm{gd}}(Y_{{\rm{fine}}}),
\label{eq:decoder}
\end{equation}
where $F_{\rm{gd}}$ is the detailed point cloud generation network.



\begin{figure}[t]   
	\centering	
	\includegraphics[width=0.85\linewidth]{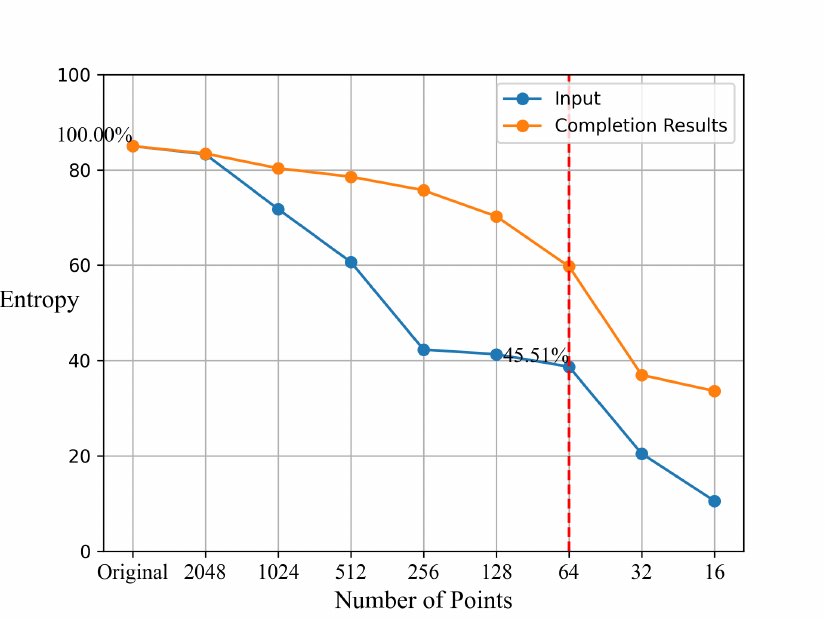}
 	\caption{We analyze the FPFH Shannon Entropy of the input point cloud and completion results on ShapeNet. We observe that when the number of input points drops to 64, the input point cloud still contains 45.51\% of the amount of information compared to the original 16,384 points (\textcolor{blue}{blue} line). When the number of points further decreases, the amount of information drops sharply. In the experiment of this paper, we use 64 as the minimum number of points. The \textcolor{orange}{orange} line shows the information of the point cloud after completion.}	
	\label{fig:FPFH_Entropy}
\end{figure}

\subsection{Training Losses}

The loss function is used to measure the difference between the output point cloud and the true point cloud. Similar to \cite{yuan2018pcn}, this loss function $L$ is defined as follows:
\begin{equation}
\begin{split}
    &L\left( {{Y_{{\rm{coarse}}}},{Y_{{\rm{detail}}}},{Y_{{\rm{gt}}}}} \right) = \\
    &{d_1}\left( {{Y_{{\rm{coarse}}}},{{\tilde Y}_{{\rm{gt}}}}} \right) + \alpha {d_2}\left( {{Y_{{\rm{detail}}}},{Y_{{\rm{gt}}}}} \right)
\label{XX}
\end{split}
\end{equation}
The above loss function consists of $d_1$, $d_2$, and weights $\alpha$. 
$d_1$ and $d_2$ represent the distance between the output point cloud and the true value at the coarse level and the detail level, respectively.

To ensure that the final output point cloud matches the true value in terms of both density distribution and overall structure, while keeping the computational effort low, we utilize two methods in our implementation. We use Earth Mover's Distance (EMD)~\cite{rubner2000earth} for ${d_1}$ calculation, which involves fewer points but provides high-density consistency with high computational complexity. For ${d_2}$ calculation, we use Chamfer Distance (CD)~\cite{fan2017point}, which involves more points but provides high structural consistency with low computational complexity. Specifically,

\small
\begin{equation}
\begin{split}
&{d_1}\left( {{Y_{{\rm{coarse}}}},{{\tilde Y}_{{\rm{gt}}}}} \right) = {\rm{EMD}}\left( {{Y_{{\rm{coarse}}}},{{\tilde Y}_{{\rm{gt}}}}} \right)\\
 &= \mathop {\min }\limits_{\phi :{Y_{{\rm{coarse}}}} \to {{\tilde Y}_{{\rm{gt}}}}} \frac{1}{{\left| {{Y_{{\rm{coarse}}}}} \right|}}\sum\limits_{p \in {Y_{{\rm{coarse}}}}} {{{\left\| {p - \phi \left( p \right)} \right\|}_2}}.
\end{split}
\label{eq:d1}
\end{equation}
\normalsize
The above equation finds the bijection $\phi$ from ${Y_{{\rm{coarse}}}}$ to ${\tilde Y_{{\rm{gt}}}}$ such that the average distance between the output point and the corresponding real point is minimized.
\begin{footnotesize}
\begin{equation}
\begin{split}
&{d_2}\left( {{Y_{{\rm{detail}}}},{Y_{{\rm{gt}}}}} \right) = {\rm{CD}}\left( {{Y_{{\rm{detail}}}},{Y_{{\rm{gt}}}}} \right)\\
 &= \frac{1}{{\left| {{Y_{{\rm{detail}}}}} \right|}}\sum\limits_{p \in {Y_{{\rm{detail}}}}} {\mathop {\min }\limits_{q \in {Y_{{\rm{gt}}}}} {{\left\| {p - q} \right\|}_2}} {\rm{ + }} 
\frac{1}{{\left| {{Y_{{\rm{gt}}}}} \right|}}\sum\limits_{q \in {Y_{{\rm{gt}}}}} {\mathop {\min }\limits_{p \in {Y_{{\rm{detail}}}}} {{\left\| {q - p} \right\|}_2}}.
\end{split}
\label{eq:d2}
\end{equation}
\end{footnotesize}
The above symmetric form of the chamfer distance is used to calculate the average closest distance between the output point cloud and the real point cloud to ensure that they are covered by each other. More training details can be found in Supplementary Material Sec. 1.

\begin{table*}[t]
\centering
\resizebox{0.74\linewidth}{!}{
\begin{tabular}{l|c|cccccccc}
\hline
              & Avg CD-${\ell _1}\downarrow$        & Airplane              & Cabinet               & Car               & Chair               & Lamp               & Sofa               & Table               & Vassel               \\
\hline
GRNet~\cite{xie2020grnet}           & 17.61           & 19.40          & 23.32           & 14.62           & 10.54           & 15.64           & 16.39           & 16.71           & 24.26           \\
PoinTr~\cite{yu2021pointr}            & 13.68           & 10.53          & 15.72           & 12.59           & 13.84           & 14.59           & 15.45           & 14.70           & 11.98           \\
PCN~\cite{yuan2018pcn}            & 12.11           & 14.53          & 11.50           & 13.32           & \textbf{6.90}          & 12.01           & 11.37           & 14.44           & 12.80           \\
SeedFormer~\cite{zhou2022seedformer}&12.38 &8.56 &16.14 &11.38 &14.25&10.11&17.50 &11.50 &9.62\\
SVDFormer~\cite{Zhu2023svdformer}& 11.22 &7.21 &13.86 &11.00 & 13.15& \textbf{10.02}& 13.90& 10.64&10.02\\
\hline
\textbf{Ours } & \textbf{7.89 } & \textbf{3.47 } & \textbf{8.93 } & \textbf{6.70 } & 10.12 & 10.31 & \textbf{10.83 } & \textbf{7.61 } & \textbf{7.63 } \\
\hline
\end{tabular}
}
\caption{64 points completion accuarry on the 8 ``seen categories" of ShapeNet. We can see that our method outperforms previous methods on average and in most categories. Avg CD-${\ell _1}$ the lower the better. \textbf{Bold} number means the best. }
\label{tab:seen}
\end{table*}


\begin{table*}[t]
\begin{minipage}[h]{0.74\linewidth}
\centering
\resizebox{1\linewidth}{!}{
\begin{tabular}{l|c|cccccccc}
\hline
& Avg CD-${\ell _1} \downarrow$              & Airplane       & Cabinet        & Car            & Chair          & Lamp           & Sofa           & Table          & Vessel \\
\hline
3D-EPN~\cite{dai2017shape}        & 20.15         & 13.16         & 21.80         & 20.31         & 18.81         & 25.75         & 21.09         & 21.712         & 18.54         \\
FoldingNet~\cite{yang2018Foldingnet}     & 14.31          & 9.49           & 15.80           & 12.61          & 15.55          & 16.41          & 15.97          & 13.65          & 14.99          \\
AtlasNet~\cite{Groueix2020AtlasNet}        & 10.85          & 6.37           & 11.94          & 10.10           & 12.06          & 12.37          & 12.99          & 10.33          & 10.61          \\
CRN~\cite{wang2020cascaded}             & 11.98          & 6.44          & 15.09         & 13.75         & 12.37         & 11.38          & 14.70         & 11.78         & 10.33         \\
MSN~\cite{liu2020morphing}           & 10.00             & 5.60            & 11.90           & 10.30           & 10.20           & 10.70           & 11.60           & 9.60            & 9.90            \\
PCN~\cite{yuan2018pcn}            & 9.64          & 5.50          & 10.63         & 8.70          & 11.00         & 11.34         & 11.68         & 8.59           & 9.67          \\
TopNet~\cite{tchapmi2019topnet}        & 9.89           & 6.24          & 11.63         & 9.83          & 11.50         & 9.37          & 12.35         & 9.36          & 8.85          \\
GRNet~\cite{xie2020grnet}         & 8.83          & 6.45           & 10.37         & 9.45          & 9.41          & 7.96          & 10.51         & 8.44          & 8.04          \\
PMP-Net~\cite{wen2021pmp}       & 8.66           & 5.50            & 11.10           & 9.62           & 9.47           & 6.89           & 10.74          & 8.77           & 7.19           \\
PoinTr~\cite{yu2021pointr}         & 8.38           & 4.75           & 10.47          & 8.68           & 9.39           & 7.75           & 10.93          & 7.78           & 7.29           \\
SCRN~\cite{wang2021cascaded}          & 8.29          & 4.80          & 9.94          & 9.31          & 8.78          & 8.66          & 9.74          & 7.20          & 7.91          \\
NSFA~\cite{zhang2020detail}          & 8.06           & 4.76           & 10.18          & 8.63           & 8.53           & 7.03           & 10.53          & 7.35           & 7.48           \\
SnowFlake~\cite{xiang2021snowflakenet}     & 7.21           & 4.29           & 9.16           & \textbf{8.08 } & 7.89           & 6.07           & \textbf{9.23 } & 6.55           & 6.40            \\
\hline
\textbf{Ours } & \textbf{7.02 } & \textbf{4.07 } & \textbf{9.12 } & 8.10            & \textbf{7.21 } & \textbf{5.88 } & 9.30            & \textbf{6.26 } & \textbf{6.25 } \\
\hline
\end{tabular}
}
\caption{2048 points completion accuarry on the 8 ``seen categories" of ShapeNet. We can see that our method outperforms previous methods in most categories and the average results.} 
\label{tab:sota}
\end{minipage}
\begin{minipage}[h]{0.245\linewidth}
\centering
\resizebox{1\linewidth}{!}{
\begin{tabular}{l|c}
\hline
Methods & Avg CD-${\ell _2} \downarrow$ \\
\hline
PCN~\cite{yuan2018pcn} & 18.22 \\
AtlasNet~\cite{Groueix2020AtlasNet} & 17.77 \\
TopNet~\cite{tchapmi2019topnet} & 14.25 \\
SoftPoolNet~\cite{Wang2020SoftPoolNet} & 11.07 \\
GRNet~\cite{xie2020grnet} & 10.64 \\
PMP-Net~\cite{wen2021pmp} & 9.23 \\
PoinTr~\cite{yu2021pointr} & 9.22 \\
CRN~\cite{wang2020cascaded} & 9.21 \\
SCRN~\cite{wang2021cascaded} & 9.13 \\
VRCNet~\cite{pan2021vrcnet} & 8.12 \\
PMP-Net++~\cite{wen2022pmpnetplusplus} & 7.97\\
Snowflake~\cite{xiang2021snowflakenet} & 7.60\\
ASFM-Net~\cite{xia2021asfm} & 6.68\\
\hline
\textbf{Ours} & \textbf{6.66}\\
\hline
\end{tabular}
}
\caption{2048 points comletion accuracy on Completion3D.}
\label{tab:com3d}
\end{minipage}

\end{table*}

\begin{table*}[t]
\centering
\resizebox{\linewidth}{!}{
\begin{tabular}{l|ccccccccccc|c}
\hline
Methods & PCN~\cite{yuan2018pcn}    & NSFA~\cite{zhang2020detail}  & CRN~\cite{wang2020cascaded} & PFNet~\cite{huang2020pfnet} & GRNet~\cite{xie2020grnet} & PoinTr~\cite{yu2021pointr} & SeedFormer~\cite{zhou2022seedformer} &AnchorFormer~\cite{Chen2023AnchorFormer} &AdaPoinTr~\cite{yu2023adapointr} &\textbf{Ours}\\
\hline
MMD↓ & 1.366  & 0.891 & 0.872 & 0.792 &  0.568 & 0.526 & 0.516 & 0.458 & 0.392 & \textbf{0.239} \\
\hline
\end{tabular}
}
\caption{Completion results on KITTI dataset with various number of input points.}
\vspace{-0.5cm}
\label{tab:KITTI1}
\end{table*}

\section{Experiments}

\subsection{Data Generation and Datasets}
\label{sec:data}

\textbf{ShapeNet data generation.} We train and evaluate our method on ShapeNet~\cite{chang2015shapenet}. It is a synthetic CAD dataset comprising 30,974 complete and detailed 3D point clouds. There are in total 16 categories of point clouds in ShapeNet. We use 8 of them for training and testing (``seen categories"), and use the other 8 categories (``unseen categories") for the generalizability test. To simulate realistic input point clouds, we sample input point clouds from the mesh surface. Specifically, we first uniformly sample 16,384 points as Ground truth. Then, we generated partial point clouds with 2048 points by back-projecting 2.5D depth images into 3D. Subsequently, we downsampled these to create low-resolution partial point clouds with varying densities: 1024, 512, 256, 128, and 64 points. The entire dataset of 30,974 3D point clouds was divided into training (28,974 point clouds), validation (800 point clouds), and testing sets (1,200 point clouds). We use the same (\textit{CD}-${\ell _1}$) as in PCN~\cite{yuan2018pcn} to evaluate our completion accuracy on both datasets. \textit{CD}-${\ell _1}$ is the Chamfer Distance using L1-norm, which is defined as 
\begin{equation}
    \textit{CD-}\ell _1(\hat{P},P) = \frac{1}{|\hat{P}|} \sum_{x \in \hat{P}} \min_{y \in P} \|x - y\|_1 + \frac{1}{|P|} \sum_{y \in P} \min_{x \in \hat{P}} \|x - y\|_1,
\end{equation}
where $\hat{P}$ and $P$ are the completion and ground truth point clouds. $x$ and $y$ are points in $\hat{P}$ and $P$, respectively. $|\cdot|$ is the size of a set, and $\|\cdot\|_1$ indicates the L1-norm. To be consistent with PCN, the number reported was multiplied by 1000 in experiments.

\noindent \textbf{KITTI dataset.} The KITTI~\cite{kitti} dataset is an in-the-wild benchmark dataset extensively used in autonomous driving and computer vision research, with a strong focus on LiDAR-based point clouds. It features a comprehensive collection of point cloud data captured from LiDAR sensors, providing detailed 3D representations of urban environments for vehicle and pedestrian detection, and scene understanding tasks. We evaluate our method on this dataset using Minimum Matching Distance (MMD) in~\cite{yu2021pointr} as the metric. MMD is defined as
\begin{equation}
{\rm{MMD}}\left( {\hat P,{\bf{C}}} \right) = {\min _{P \in {\bf{C}}}}{\textit{CD-}}\ell _2\left( {\hat P,P} \right),
\end{equation}
\textit{CD}-${\ell _2}$ is the Chamfer Distance the output ${\hat P}$ and the point cloud ${P}$ from car category ${\bf{C}}$ in ShapeNet that is closest to ${\hat P}$ in terms of \textit{CD}-${\ell _2}$, where
\begin{equation}
    \textit{CD-}\ell _2(\hat{P},P) = \frac{1}{|\hat{P}|} \sum_{x \in \hat{P}} \min_{y \in P} \|x - y\|_2^2 + \frac{1}{|P|} \sum_{y \in P} \min_{x \in \hat{P}} \|x - y\|_2^2,
    \label{eq:cdl2}
\end{equation}
To be consistent with \cite{yu2021pointr}, the number reported was multiplied by 1000 in experiments.\

\noindent \textbf{Completion3D dataset.} We also compare our results on Completion3D~\cite{tchapmi2019topnet}, which is another 3D object point cloud completion benchmark designed based on the data in ShapeNet. Here, we use the same \textit{CD}-$\ell _2$ metric as defined in \cref{eq:cdl2}.

\subsection{Shannon Information Analysis}
We analyze the Shannon information for few-point inputs and completion results on ShapeNet.  We use \cref{eq:fpfh} in \cref{sec:fpfh} to evaluate the FPFH Entropy of input point clouds. We set the voxel size of FPFH to approximately 2\% of the point cloud's overall size, meaning we do not take into consideration the errors that are smaller than that. The radius and bin count are set to 1cm and 36, respectively. The results are shown in \cref{fig:FPFH_Entropy}. The \textcolor{blue}{blue} line shows the entropy of different inputs and the \textcolor{orange}{orange} line shows the results of the corresponding completed point cloud. Compared with the ground truth point cloud, our 64-point input has 45.51\% information of the ground truth
We can also see from the \textcolor{orange}{orange} line that even with 64-point input, we can also recover 60\% of the shape information.


\begin{table}

\end{table}


\begin{table*}[t]
\centering
\resizebox{0.74\linewidth}{!}{
\begin{tabular}{l|c|cccccccc}
\hline
      & Avg CD-${\ell_1}\downarrow$         & Bus              & Bed               & Bookshelf         & Bench             & Guitar            & Motorbike         & Skateboard        & Pistol            \\
\hline
GRNet~\cite{xie2020grnet}           & 17.03           & 17.59          & 25.61           & 21.93           & 15.26           & 10.85           & 14.83           & 14.96           & 15.17           \\
PCN~\cite{yuan2018pcn}           & 15.52           & 11.30          & 25.86           & 16.22           & 12.55           & 11.47           & 15.30           & 13.48            & 17.95           \\
PoinTr~\cite{yu2021pointr}            & 14.92           & 14.62          & \textbf{20.44}           & 16.84           & 14.22           & 11.56           & 13.40           & 14.45          & \textbf{13.80}          \\
SVDFormer~\cite{Zhu2023svdformer}            & 13.60           & 13.30          & 21.47           & 13.73           & 11.69           & \textbf{8.00}           & \textbf{13.25}           & \textbf{9.91}          & 17.44         \\
\hline
\textbf{Ours } & \textbf{12.90 } & \textbf{7.94 } & 22.60  & \textbf{12.98 } & \textbf{9.59 } & 9.77  & 13.56 & 11.55  & 14.92  \\
\hline
\end{tabular}
}
\caption{Results for 64 points on shapes from unseen categories in ShapeNet. We observe that even on unseen categories, we can still have reasonable results, and our method still outperforms previous methods on average and most unseen categories.}
\label{tab:unseen}
\end{table*}


\begin{figure}[t]   
	\centering	
	\includegraphics[width=0.9\linewidth]{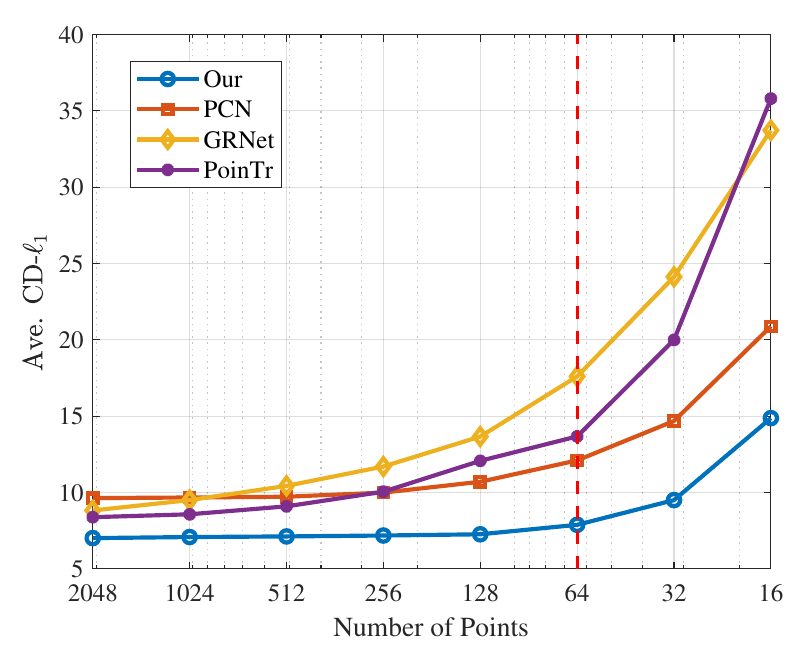}
	\caption{In our comparative analysis on ShapeNet, we varied the input point number from 2048 to 16. We observe that our model (\textcolor{blue}{blue} line) consistently outperformed previous methods across different input point counts. Notably, when the number of input points was reduced to below 64, there was a significant decline in performance for all evaluated methods.}	
	\label{fig:sn_vary}
\end{figure}

\begin{figure}[t]   
	\centering	
	\includegraphics[width=0.9\linewidth]{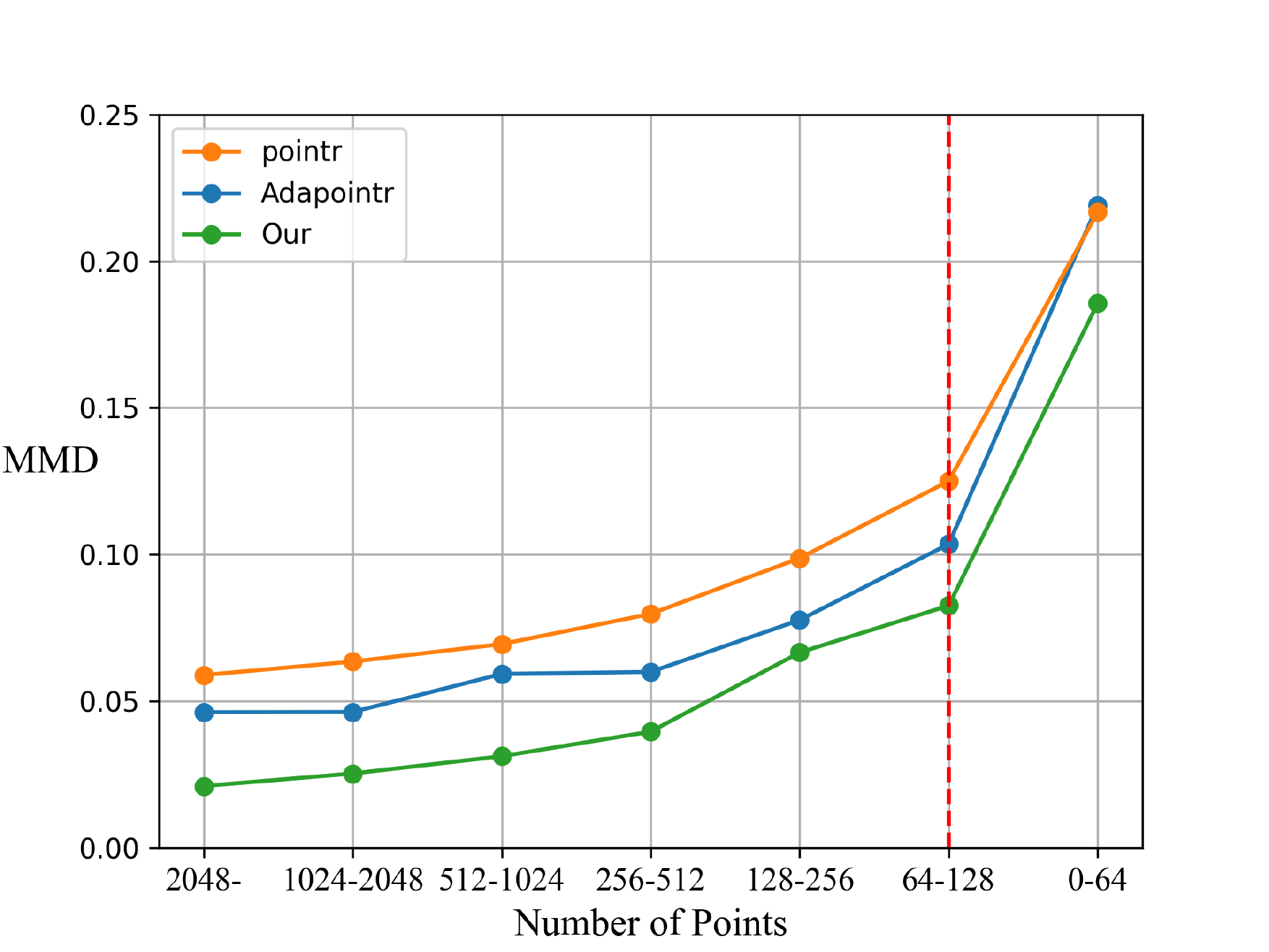}
	\caption{We report the completion result grouped by the number of input points on LiDAR scans of the KITTI dataset. As shown in the figure, our method (\textcolor{green}{green} line) outperforms previous methods in every number of input point groups. We also observe a sharp error increase when the number of input points is lower than 64.}	
	\label{fig:KITTI_MMD}
\end{figure}

\begin{figure}[t]
    \centering	
    \includegraphics[width=1\linewidth]{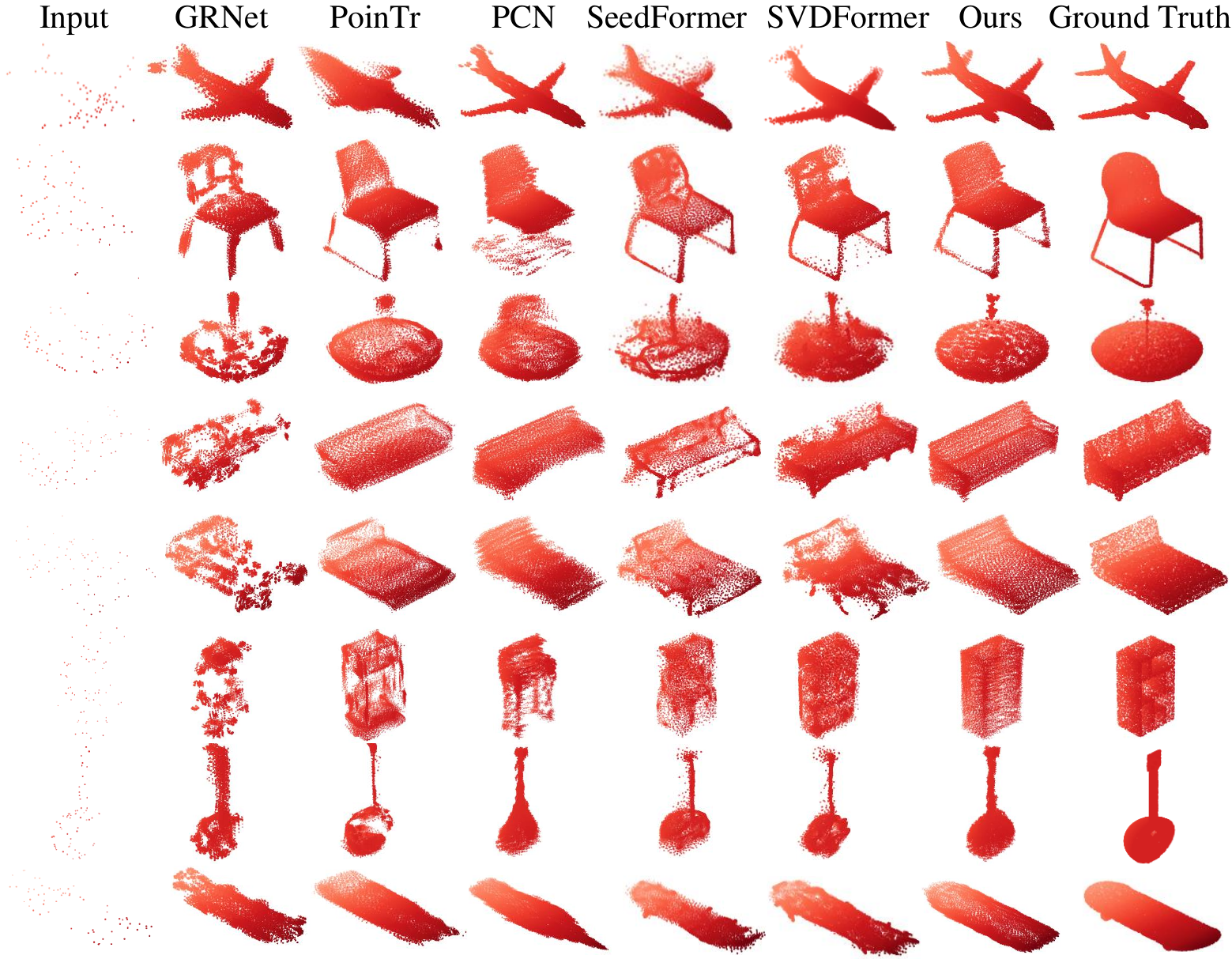}
    \caption{Visualized results for 64 input points. The top four rows are the results of categories used during training (``seen" categories). The bottom four rows are the results of categories not seen during training (``unseen" categories). As we can see, our method has better visualization results than previous methods.
    }
    \label{fig:vis_few}    
\end{figure}

\begin{figure}[t]
    \centering
    \includegraphics[width=0.85\linewidth]{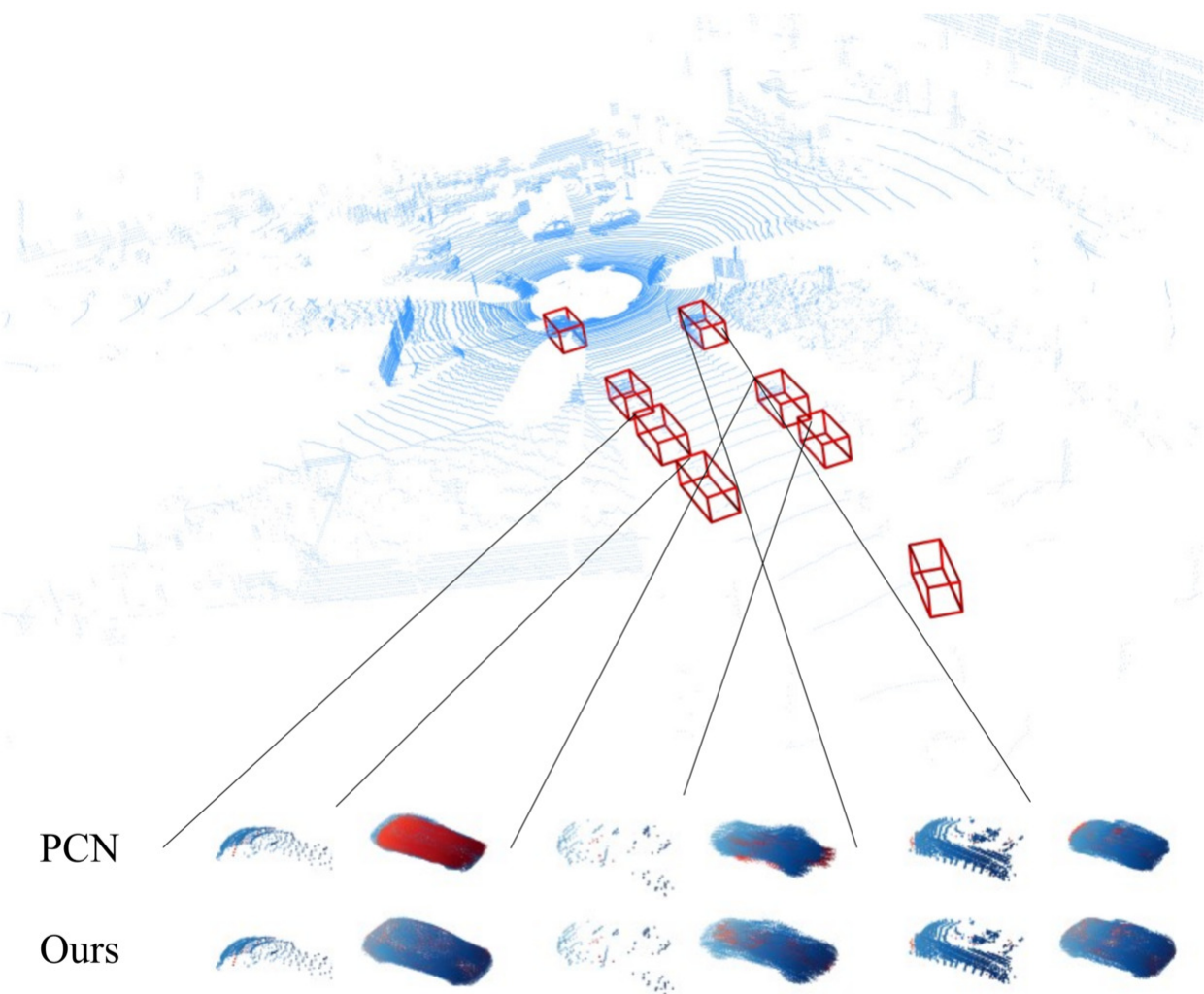}
    \caption{Visualization results compared on KITTI, Red and Blue are before after aligning the centroids of the point clouds.}	
    \label{fig1}
    \label{fig:kitti_vis}
\end{figure}

\begin{table*}[t]
\centering
\resizebox{1.0\linewidth}{!}{
\begin{tabular}{c|ccccccc|ccccccc}
\hline & \multicolumn{7}{c|}{Module} & \multicolumn{6}{c}{Number of input points}    \\ \cline{2-14}
& \makecell{Extensive branch \\ Baseline Encoder} & \makecell{Salience \\ branch} & \makecell{Feature \\ revision} & \makecell{Point cloud \\ revision} & PointNet++ & Transformer & \makecell{Decoder \\ Baseline Decoder} &2048 & 1024 & 512  & 256   & 128    & 64    \\ \hline
a& \checkmark & & & & & & \checkmark & 9.64           & 9.68          & 9.73          & 10.00          & 10.71           & 12.11            \\

b& & \checkmark & & & &   & \checkmark & 8.66           & 8.71          & 8.76          & 8.95          & 9.42           & 10.53             \\

c& \checkmark & \checkmark & & & & & \checkmark & 8.31             & 8.35           & 8.42          & 8.49           & 8.59           & 9.78            \\

d& \checkmark & & \checkmark & & & & \checkmark & 8.33           & 8.39           & 8.43           & 8.60          & 9           & 10            \\

e&\checkmark & & & \checkmark & & & \checkmark & 9.01		 & 9.11           & 9.18           & 9.27          & 10.07            & 11.35            \\

f& \checkmark & & \checkmark & \checkmark & & & \checkmark & 8.01			 & 8.07          & 8.15          & 8.29          & 8.55           & 9.14            \\

g& \checkmark & & & & \checkmark & & \checkmark & 9.31           & 9.36          & 9.41          & 9.65          & 10.28           & 11.58           \\

h& \checkmark & & & & & \checkmark & \checkmark & 8.91           & 8.99          & 9.08          & 9.17          & 9.72           & 11.02            \\

i& \checkmark & & & & \checkmark & \checkmark & \checkmark & 8.5        & 8.58           & 8.66          & 9.02           & 9.43          & 11.24    \\        

j& \checkmark & & \checkmark & \checkmark & \checkmark & \checkmark & \checkmark   & 7.97           & 8.03          & 8.07          & 8.22          & 8.51          & 9.46             \\

k& & \checkmark & \checkmark & \checkmark & \checkmark & \checkmark & \checkmark & 7.35           & 7.41           & 7.46          & 7.54           & 7.7          & 8.42    \\     

\hline
\textbf{Ours} & \checkmark &\checkmark & \checkmark& \checkmark & \checkmark &  \checkmark &  \checkmark & \textbf{7.02 } & \textbf{7.09 } & \textbf{7.13 } & \textbf{7.19 } & \textbf{7.27 } & \textbf{7.89 }\\
\hline
\end{tabular}
}
\caption{Ablation study under different numbers of input points. Our performance decreases when we remove some necessary modules.}

\label{tab:ablation_in_decrease_point}
\end{table*}

\subsection{Results}\label{sec:results}

\textbf{Few-point results on ShapeNet.} We train and evaluate our method on 8 categories of point clouds on ShapeNet. \cref{tab:seen} presents the per-class completion results on ShapeNet for scenarios where the input point cloud contains only 64 points. In all eight ``seen categories", our method outperforms others, achieving the highest accuracy. We observe a notable improvement over the previous method. This demonstrates that our approach has a distinct advantage in completion accuracy, especially in cases of few-point point cloud input. We also compare the performance on ShapeNet when the input point number varies from 2048 to 16 in \cref{fig:sn_vary}. We can observe that our \model{} outperforms previous approaches when the number of input points changes. Besides, when the number input point is less than 64, the performance of all methods drops sharply.

\noindent \textbf{Many-point results.} In \cref{tab:sota}, we present the per-class completion results for scenarios with 2048 input points on ShapeNet, and in \cref{tab:com3d} on Completion3D. These results indicate that our method also outperforms previous approaches in cases of regular input point counts (2048).

\noindent \textbf{Results on KITTI.} We directly evaluate KITTI using the model trained on ShapeNet. As a real-world dataset, the input point number of KITTI varies from 1 to 2048. We first report the overall result on KITTI regardless of the input point number in \cref{tab:KITTI1}. Our completion error is smaller than previous approaches. We also group the input point cloud in KITTI by the number of points to measure the performance when the input number of points varies. Our result is shown in \cref{fig:KITTI_MMD}. We can observe that our method outperforms previous approaches when the number of points drops. The experiments on KITTI show the generalizability of our method to different datasets.

\noindent \textbf{Results on unseen categories of ShapeNet.} \cref{tab:unseen} gives the per-class completion results for the shape from unseen categories when the input point cloud only contains 64 points. Our method still achieves the best performance in all 8 unknown categories. This experiment shows our method has reasonable generalizability to unseen objects.

\begin{figure}[t]
    \centering	
    \includegraphics[height=0.17\linewidth]{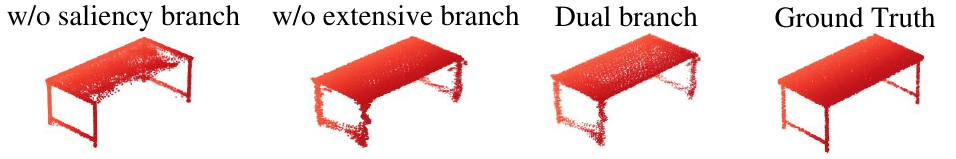}
    \caption{Visualized comparison of removing each branch in feature extractor. From left to right: removing saliency branch, removing extensive branch, keeping both branch (proposed method), and Ground Truth.}
    \label{fig:abla_branch}
\end{figure}

\noindent \textbf{Ablation studies.} 
\cref{tab:ablation_in_decrease_point}
gives the contribution of each module to the performance improvement under different numbers of input points, respectively. No matter how many input points there are and what we choose, using the salient branch instead of the extensive branch can always improve the completion accuracy, and by combining the two branches together, the accuracy can be further improved. This justifies our design of dual-branch feature extraction and the two-stage revision network. We also show an visualized example of removing one of feature extraction dual branches in \cref{fig:abla_branch}. Our proposed method (third one from the left) achieves the best results.

\noindent \textbf{Visualizations.} \cref{fig:vis_few} shows visualized results on ShapeNet. The input point cloud only contains 64 points. As we can see, GRNet~\cite{xie2020grnet} may only generate some non-connected clusters around isolated input points, whereas the category information may be completely lost. Although PCN~\cite{yuan2018pcn} can generate connectivity, the results often contain large deformation and noise, leading to either difficulty or error in recognition. Although our method also contains deformation and noise, most of them are within an acceptable range. This illustrates the effectiveness of our method for the very low-resolution case. Besides, our visualized result on KITTI shows a similar result in \cref{fig:kitti_vis}.

\begin{figure}[t]
    \centering	
    \includegraphics[height=0.28\linewidth]{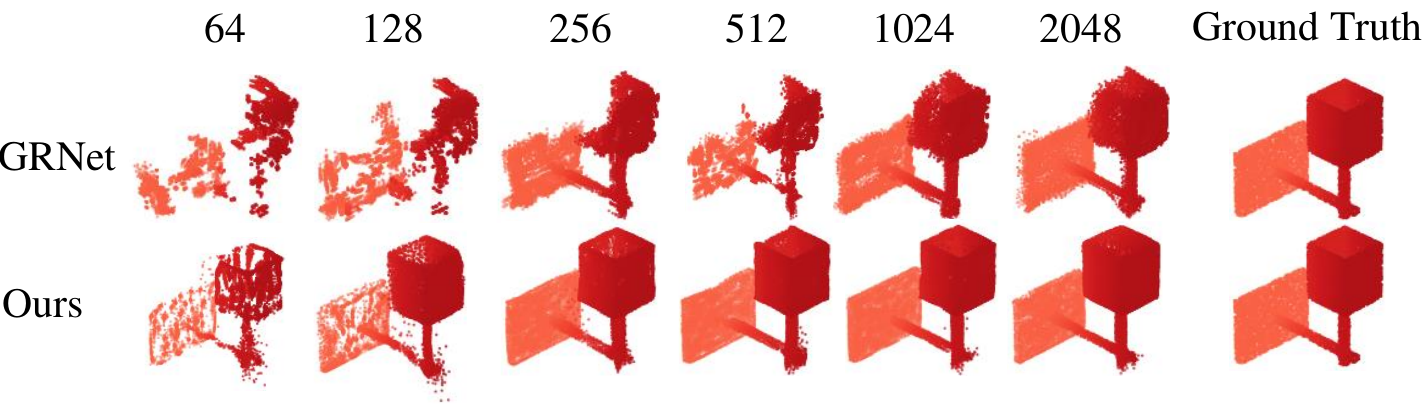}
    \caption{Qualitative completion results of a lamp when the number of points increases from 64 to 2048. Our visualization results outperform GRNet for different numbers of points.}
    \label{fig:airplane}
\end{figure}

\noindent \textbf{Visualized result when input point number changes.} \cref{fig:airplane} gives the completion results of an airplane when the number of input points gradually decreases from 2048 to 16. As the number of input points decreases, our method always recovers a better point cloud, which further illustrates our stability against the variations in the number of points of the input.

\section{Conclusion}

We have proposed a point cloud complementation network specifically for the case of very few points. To overcome the problem of missing neighborhood information due to very few points, we select a PCN that does not rely on neighborhood information in the encoding phase as the backbone network and combine a series of methods applicable to the very few points cases to enhance the overall performance of the complementation network. These specialized designs include 1) a dual-branch feature extraction network, 2) a two-stage revision network, and 3) a detailed point generation network with fused point features. Experimental results show that our method has a significant performance improvement over existing methods in very few point cases. Also, our method has good robustness for the shapes from unseen categories and the number of different points.

{
    \small
    \bibliographystyle{ieeenat_fullname}
    \bibliography{main}
}


\end{document}